\def\year{2020}\relax
\documentclass[letterpaper]{article} % DO NOT CHANGE THIS
\usepackage{aaai20}  % DO NOT CHANGE THIS
\usepackage{times}  % DO NOT CHANGE THIS
\usepackage{helvet} % DO NOT CHANGE THIS
\usepackage{courier}  % DO NOT CHANGE THIS
\usepackage[hyphens]{url}  % DO NOT CHANGE THIS
\usepackage{graphicx} % DO NOT CHANGE THIS
\usepackage{graphicx}  % DO NOT CHANGE THIS
\title{AAAI Press Formatting Instructions \\for Authors Using \LaTeX{} --- A Guide }
\usepackage{times}
\usepackage{latexsym}
\usepackage{booktabs}
\usepackage{helvet}  %Required
\usepackage{courier}  %Required
\usepackage{graphicx}  %Required
\usepackage{color, xcolor}
\usepackage{multirow}
\usepackage{multicol}
\usepackage{amsmath}
\usepackage{amssymb}
\usepackage{latexsym}
\usepackage{textcomp}
\usepackage{subfig}
\usepackage{array}
\usepackage{xspace}
\usepackage{wasysym}
\usepackage{color}
\usepackage{url}
\usepackage{sidecap}
\usepackage{arydshln}

\newcommand{\newcite}[1]{\citeauthor{#1}~(\citeyear{#1})}

\DeclareMathOperator*{\argmax}{arg\,max}

\definecolor{forestgreen}{rgb}{0.0, 0.50, 0.0}
\definecolor{goldenbrown}{rgb}{0.6, 0.4, 0.08}

\title{Go From the General to the Particular: Multi-Domain Translation\\ with Domain Transformation Networks}

\author{Yong Wang\thanks{Work done when interning at Tencent AI Lab.} \\ The University of Hong Kong \\ {wangyong@eee.hku.hk}   \And
Longyue Wang \\ Tencent AI Lab \\ {vinnylywang@tencent.com} \AND
Shuming Shi \\ Tencent AI Lab \\ {shumingshi@tencent.com} \And
Victor O.K. Li \\ The University of Hong Kong \\ {vli@eee.hku.hk} \And
Zhaopeng Tu \\ Tencent AI Lab \\ {zptu@tencent.com}
}

\date{}

\begin{document}
\maketitle

\begin{abstract}
The key challenge of multi-domain translation lies in simultaneously encoding both the general knowledge shared across domains and the particular knowledge distinctive to each domain in a unified model. Previous work shows that the standard neural machine translation (NMT) model, trained on mixed-domain data, generally captures the general knowledge, but misses the domain-specific knowledge. In response to this problem, we augment NMT model with additional {\em domain transformation networks} to transform the general representations to domain-specific representations, which are subsequently fed to the NMT decoder. To guarantee the knowledge transformation, we also propose two complementary supervision signals by leveraging the power of knowledge distillation and adversarial learning. Experimental results on several language pairs, covering both balanced and unbalanced multi-domain translation, demonstrate the effectiveness and universality of the proposed approach. Encouragingly, the proposed unified model achieves comparable results with the fine-tuning approach that requires multiple models to preserve the particular knowledge. Further analyses reveal that the domain transformation networks successfully capture the domain-specific knowledge as expected.\footnote{The source code and experimental data are available at \url{https://github.com/wangyong1122/dtn}.}
\end{abstract}

\section{Introduction}
\label{sec:intro}

In multi-domain translation, a unified neural machine translation (NMT) model is expected to provide high quality translations across a wide range of diverse domains. The main challenge of multi-domain translation lies in learning a unified model that simultaneously 1) exploits the {\em general knowledge} shared across domains, and 2) preserves the {\em particular knowledge} that represents distinctive characteristics of each domain.
Unfortunately, standard NMT models trained on the mixed-domain data generally capture the general knowledge while ignoring the particular knowledge, rendering them sub-optimal for multi-domain translation~\cite{koehn2017six}.

A natural approach to this problem is fine-tuning, which first trains a general model on all data and then separately fine-tunes it on each domain~\cite{luong2015stanford}. However, the fine-tuning approach requires maintaining a distinct NMT model for each domain, which makes it unwieldy in practice. Towards learning a unified multi-domain translation model, several researchers turn to augment the NMT model to learn domain-specific knowledge. For example,~\newcite{kobus2016domain} introduced a special domain tag to the source sentence, and~\newcite{britz2017effective} and~\newcite{zeng2018multi} guide the encoder output to embed domain-specific knowledge via an auxiliary object. 
However, all the approaches require the encoder representations to embed both the general and the particular knowledge at the same time.
Recent studies have shown that such overloaded usage of hidden representation makes training the model difficult, and such problem can be mitigated by separating these functions~\cite{Rocktaschel:2017:ICLR,Zheng:2018:TACL}.

In this work, we explicitly model the domain-specific functionality for multi-domain translation by introducing {\em domain transformation networks} (\textsc{Dtn}s). 
More specifically, the \textsc{Dtn}s transform the general knowledge learned by the encoder to the domain-specific knowledge, which is subsequently fed to the decoder. 
In this way, the encoder learns general knowledge in the standard fashion, and the newly added \textsc{Dtn}s learn to preserve the particular knowledge. 
We employ a residual connection on \textsc{Dtn}s to enable the decoder to exploit both the general and particular knowledge.
To guarantee the knowledge transformation, we also propose two supervision strategies: 
1) {\em domain distillation} that encourages the unified model to learn domain-specific knowledge in a teacher-student framework;
and 2) {\em domain discrimination} that guides the encoder output and the transformed representation to embed the required knowledge with adversarial learning. 

We conduct experiments on three language pairs: Chinese$\Rightarrow$English, German$\Rightarrow$English and English$\Rightarrow$French, covering balanced, unbalanced and large-scale multi-domain data. 
Experimental results show that our model significantly and consistently outperforms both the \textsc{Transformer} baseline by +3.35 BLEU points and previous multi-domain translation models~\cite{kobus2016domain,britz2017effective,zeng2018multi} by +1.0$\sim$2.0 BLEU points on different data, demonstrating the effectiveness and universality of the proposed approach.
Encouragingly, our unified model is on par with the fine-tuning approach that requires multiple models to preserve the particular knowledge.
Further analysis reveals that the domain transformation networks successfully capture the domain-specific knowledge while maintaining the specificity of each domain.

\paragraph{Contributions}
Our main contributions are:
\begin{enumerate}
\item Our study demonstrates the necessity of explicitly modeling the transformation from the general to the particular for multi-domain translation.

\item We exploit two supervision signals to simultaneously and incrementally encourage transformation of domain knowledge.

\item We construct several multi-domain data across languages, on which we empirically validate a variety of existing approaches.
\end{enumerate}

\section{Background}
\paragraph{Neural Machine Translation}
A standard NMT model directly optimizes the conditional probability of a target sentence $\mathbf{y} = y_1, \dots, y_{J}$ given its corresponding source sentence $\mathbf{x} = x_1, \dots, x_{I}$:
\begin{equation}
P(\mathbf{y}|\mathbf{x}; \theta) = \prod_{j=1}^{J} P(y_j | \mathbf{y}_{<j}, \mathbf{x}; \theta)
\end{equation}
where $\theta$ is a set of model parameters and $\mathbf{y}_{<j}$ denotes the partial translation. The probability $P(\mathbf{y}|\mathbf{x}; \theta)$ is defined on the neural network based encoder-decoder framework~\cite{sutskever2014sequence,cho2014learning}, where the encoder summarizes the source sentence into a sequence of representations $\mathbf{H}=\mathbf{H}_1,\dots,\mathbf{H}_I$ with $\mathbf{H} \in \mathbb{R}^{I \times {d}}$, and the decoder generates target words based on the representations. Typically, this framework can be implemented as recurrent neural network (RNN)~\cite{bahdanau2014neural}, convolutional neural network (CNN)~\cite{gehring2017convolutional} and Transformer~\cite{vaswani2017attention}. %{\color{red}In this study, we implement our approach and re-implement existing models on top of Transformer.} 
The parameters of the NMT model are trained to maximize the likelihood of a set of training examples $D=\{[\mathbf{x}^m,\mathbf{y}^m]\}_{m=1}^M$:
\begin{eqnarray}
\label{org-loss}
\mathcal{L}(\theta) = \argmax_{\theta} \sum_{m=1}^{M} \log P({\bf y}^m|{\bf x}^m; \theta)
\end{eqnarray}
The training corpus generally consists of data from various domains, which are not distinguished by the NMT model. This may pose difficulties to multi-domain translation.

\paragraph{Multi-Domain Translation}
This task aims to build a unified model on the mixed-domain data by maximizing performances across all domains. 
Formally, there are $N$ subsets $D_1,...,D_N$ from different domains, where the $n$-th domain of subset $D_n=\{[\mathbf{x}_n^m,\mathbf{y}_n^m]\}_{m=1}^{M_n}$. Accordingly, the training objective is
\begin{equation}
    \mathcal{J}(\theta) = \argmax_{\theta} \frac{1}{N} \sum_{n=1}^N \mathcal{L}_n(\theta)
\label{eq.multi-domain}
\end{equation}
which maximizes the likelihood over training examples in each domain (i.e., $\mathcal{L}_n(\theta)$). As seen, there is no explicit signals to guide the model to learn domain-aware information in the learning objective function. As a result, the parameters in a standard NMT model generally capture the general knowledge while ignoring the domain-specific knowledge.

\begin{figure*}[t]
\centering
\includegraphics[width=\textwidth]{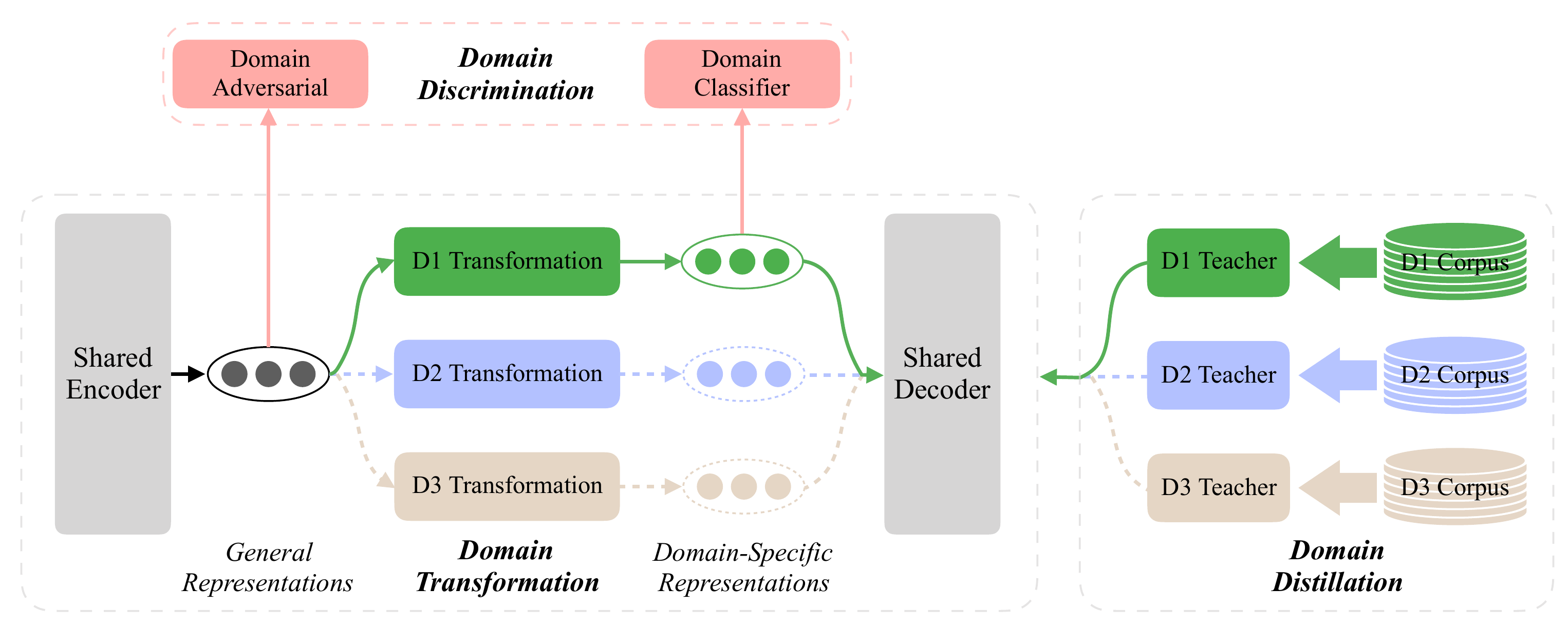}
\caption{Architecture of the proposed multi-domain translation model, which consists of two key components: 1) {\em domain transformation} that transforms from the general representations to domain-specific representations, and we maintain a distinct transformation network for each domain; 2) {\em domain supervision} that contains two sub-components: {\em domain distillation} and {\em domain discrimination}. Domain distillation learns domain-specific model guided by domain teachers, which are fine-tuned on corresponding training corpora. Domain discrimination guides the two types of representations to embed the required content. In this example, the data of Domain 1 (``D1'') are used to train the model, and solid line denotes the information flow.}
\label{fig-e2e}
\end{figure*}

\section{Approach}
\label{sec:app}

Our goal is to build a unified model, which can achieve good performance on all domains. As shown in Figure~\ref{fig-e2e}, we augment the standard NMT model with the introduced {\em Domain Transformation} networks, which transform the general encoding representations to the domain-specific representations. To guarantee the knowledge transformation effectively, we also propose two complementary supervision signals: {\em Domain Distillation} and {\em Domain Discrimination}, leveraging the power of knowledge distillation and adversarial learning. 

\subsection{Domain Transformation}
\label{sec:3.1}

\paragraph{Residual Transformation Networks}
The basic idea of domain transformation is to separate the specific features of each domain from the general features across multiple domains. First, we learn a shared encoder that maps input sentences to general representations that preserve common knowledge regardless of domains. Simultaneously, we learn a transformation component that explicitly transforms the general representations to domain-specific representation spaces, each of which represents distinctive characteristics of one single domain. The residual connection on transformation networks implicitly serves as an interpolation of the general and domain specific representations.

Formally, the transformation module reads a sequence of hidden states and outputs transformed ones. The source sentence $\mathbf{x}$ is first summarized into general representations $\mathbf{H}$ by a shared encoder of the standard NMT model. Conditioned on the input latent representations $\mathbf{H}$, we then employ a residual model~\cite{he2016deep} to generate domain-specific representations $\mathbf{H'}$ by:
\begin{equation}
\mathbf{H'}=\mathcal{F}(\mathbf{H}, W_{n})+\mathbf{H}
\end{equation}
where $W_{n}$ is the parameters related to the $n$-th domain and $\mathcal{F}(\cdot)$ is the functional mapping which can be implemented by different types of neural networks such as feed-forward network~(FNN), CNN and self-attention network~(SAN). Subsequently, the output representations $\mathbf{H'}$, which encode both the general knowledge $\mathbf{H}$ and the domain-specific knowledge $\mathcal{F}(\mathbf{H}, W_{n})$, are fed to a shared decoder for generating the target sentence $\mathbf{y}$.

The differences between domains are usually uncertain and tiny, which leads to inefficiency of directly fitting a desired underlying mapping. Recently, ``residual'' -- a concept in deep neural networks~\cite{he2016deep,sohn2019unsupervised} has been successfully applied to extract feature differences in fields of image classification~\cite{sohn2019unsupervised} and speech recognition~\cite{van2016wavenet} and achieves remarkable improvements of performance. In our preliminary experiments, we have investigated two different implementations of transformation networks including stacked feed-forward networks and multi-head attention networks. We found that the multi-head attention mechanism performs better in respect of capturing such domain-aware characteristics for the multi-domain translation task. In this study, we also parameterized $\mathcal{F}(\cdot)$ based on domain symbols, where each transformation module is able to maintain its own domain-aware parameters. 

\paragraph{Domain-Aware Batch Learning}
To train distinct parameters of each domain, we propose a domain-aware learning strategy, in which one batch only contains training examples in a certain domain. 
One straightforward implementation is to alternately or randomly feed domain-aware batches into our proposed model. 
However, in the preliminary experiments, we found severe overfitting problems when using unbalanced multi-domain data. 
To overcome this, we propose a more balanced method, which heuristically selects a certain domain-batch by considering its distribution over the entire training corpus. Formally, domain-batches are sampled according to a multinomial distribution with probabilities $\{q_i\}_{i=1,...,N}$:
\begin{equation}
q_i = \frac{p_i^\alpha}{\sum_{j=1}^Np_j^\alpha}\quad \quad p_i = \frac{n_i}{\sum_{k=1}^Nn_k}
\end{equation}
where $n_i$ is the number of batches of the $i$-th domain and $\alpha=0.7$ is the balance factor, which aims to increase the number of tokens associated with low-resource domains and alleviates the bias towards high-resource domains.

\subsection{Domain Supervision}
\label{sec:3.2}

\paragraph{Domain Distillation}
The generalization ability of the teacher model can be transferred to the student by using the class probabilities produced by the cumbersome model for training the small model~\cite{hinton2015distilling}.
Recent studies on speech recognition show that training student networks with multiple teachers achieves promising empirical results~\cite{you2019teach}. 

Inspired by these observations, we propose to teach a unified model with multiple teachers trained on different domains. Specifically, we employ the soft targets produced by fine-tuned models as the supervision signal to train our unified model with the benefits of exploiting more data information and simultaneously reducing the interference across domains.
 
For the learning objective, we linearly interpolate soft target distribution produced by the corresponding domain teacher with hard labels:
\begin{equation}
\begin{split}
\mathcal{L}(\theta)=\argmax_{\theta}\sum_{(\mathbf{x},\mathbf{y})\in D}\bigg\{(1-\lambda)\log P(\mathbf{y}|\mathbf{x};\theta)  & \\
+\lambda\sum_{j=1}^J\sum_{k=1}^{|V|}\hat{P}(y_j=k|\mathbf{y}_{<j},\mathbf{x};\hat{\theta}) & \\
\times\log P(y_j=k|\mathbf{y}_{<j},\mathbf{x};\theta)\bigg\} &
\label{eq.kd}
\end{split}
\end{equation}
where $\lambda$ is a hyper-parameter that is shared across multiple domains, $|V|$ is the vocabulary size of the target language, and $\hat{P}(\cdot)$ is the soft target.

\paragraph{Domain Discrimination}
Adversarial and discriminative learning can effectively distinguish between different types of features~\cite{ganin2015unsupervised,chen2017adversarial,sun2018domain,adams2019massively}. 
In this work, we augment the transformation networks with the ability of domain discrimination. 
Specifically, the adversarial domain classifier is deployed at the input of \textsc{Dtn}s, namely:
\begin{equation}
P(d|\mathbf{x};\psi)=\textup{softmax}(W_{D}^\top \mathbf{\widetilde{H}})
\end{equation}
where $d$ is the domain symbol, $W_{D}$ is the weights of softmax classifier and $\mathbf{\widetilde{H}}$ is the weighted representations of the encoding representations ($\mathbf{H}$), which is calculated as follows:
\begin{equation}
\mathbf{\widetilde{H}}=\sum_{i=1}^I\alpha_i\mathbf{H}_i
\label{eq.domain classifier}
\end{equation}
where the computation of $\alpha_i$ is similar to self-attention~\cite{lin2017structured}, in which the query is a trainable vector.
Furthermore, we conduct a domain classifier on the output of \textsc{Dtn}s $\mathbf{\widetilde{H}'}$ to guide it to embed domain-specific knowledge:
\begin{equation}
P(d|\mathbf{x};\gamma)= \textup{softmax}(W_{D}'^\top\mathbf{\widetilde{H}'})
\end{equation}
where $\gamma$ is a set of parameters of the domain classifier and $\mathbf{\widetilde{H}'}$ can be obtained according to Equation~\eqref{eq.domain classifier}.

Overall, the training objective is a linear interpolation of the likelihood and the domain discrimination:
\begin{equation}
\begin{split}
\mathcal{L}&(\theta, \gamma, \psi) =\argmax_{\theta, \gamma, \psi}  \sum_{(\mathbf{x},\mathbf{y},d)\in D} \\
&~~~~~~\bigg\{ \underbrace{\log P({\bf y}|{\bf x}; \theta)}_\text{\normalsize \em likelihood} + \underbrace{\log P(d|\mathbf{x};\gamma)}_\text{\normalsize \em domain classifier}  \\ 
& + \underbrace{\log P(d|\mathbf{x};\psi)+ \delta \times H(P(d|\mathbf{x};\psi))}_\text{\normalsize \em domain adversarial} \bigg\} 
\label{eq.training objective}
\end{split}
\end{equation}
where $\delta$ is the balance factor, and $H(P(\cdot))$ is the entropy of the probability distribution of the adversarial domain classifier with $N$ domain labels. 
Following \newcite{zeng2018multi} and \newcite{chen2017teacher}, we also employed the two-phase training strategy, where we alternatively optimized $\mathcal{L}(\theta,\gamma,\psi)$ with $\{\theta,\gamma\}$ and $\psi$. Besides, we discarded the component $\log P(d|\mathbf{x};\psi)$ when training the $\{\theta,\gamma\}$ parameter set.

\paragraph{Discussion}
While these two supervisions have their own characteristics, domain distillation and discrimination are complementary to each other.
Domain distillation exploits more information of data, including shared and domain-aware knowledge across domains.
As a strong supervision signal, domain discrimination is used to guide the transformation model to learn the distinct information between general and domain-specific representations.

\section{Experiments}
\subsection{Setup}

\begin{table}[t]
	\centering
	\begin{tabular}{c | c r || c | c r }
	Corpus & D & $|S|$ & Corpus & D  &  $|S|$ \\
	\hline 
	  & Law & 0.22          &       \multirow{4}{*}{De$\Rightarrow$En} & Law & 0.59          \\
	{Zh$\Rightarrow$En}  & Oral & 0.22 &      & Med. & 0.87\\
	(small) & Thesis & 0.30    &   & IT & 0.31\\
	& News & 0.30  &     & Koran & 0.53\\
	\hline
    & Law & 1.46   &   &   \\
    {Zh$\Rightarrow$En}  & News & 1.54   &   \multirow{2}{*}{En$\Rightarrow$Fr}  & Med. & 0.89\\
    (large)     & Patent & 2.90  &   &   Parl. & 2.04\\
    & Sub. & 1.77 &   &\\
	\end{tabular}
	\caption{Statistics of training corpora: ``D'' and ``$|S|$'' indicate the domain and the number of sentences (in millions). As seen, Zh$\Rightarrow$En can be regarded as ``balanced data'' as the number of training samples is similar across domains while De$\Rightarrow$En and En$\Rightarrow$Fr are ``unbalanced data'' as the numbers of sentence pairs are very different.}
	\label{tab-data}
\end{table}

\begin{table*}[t]
\centering
\begin{tabular}{c|l|c|r|llll|lc}
	\multirow{2}{*}{\bf \#}    &  \multirow{2}{*}{\bf Architecture}   &    \multirow{2}{*}{\bf \#M}       & \multirow{2}{*}{\bf \#Para.}  &  \multicolumn{6}{c}{\bf BLEU}\\
	\cline{5-10}
	    & & & & Law & Oral & Thesis & News & Avg. & {\bf  $\bigtriangleup$}\\
	\hline\hline
	\multicolumn{9}{c}{RNN-based NMT~\cite{zeng2018multi}} \\
	\hline
	1   &   RNNSearch  &    \multirow{2}{*}{1}   &  --  &   45.82    &  9.15   &  13.93 & 19.73  & 22.16 & -- \\
	2   &   ~~~ + Domain Context   &       &  --  &   55.03    &  10.20   &  18.04 & 22.29  & 26.39 & -- \\
	\hline
	\multicolumn{9}{c}{Transformer-based NMT (\em this work)} \\
	\hline
    3 &   Transformer     &   1    &  95.2M  &   65.72    &   10.28   & 20.38 & 27.22  & 30.90 & -- \\
	\hline
	4 & ~~~ + Fine-tune~\cite{luong2015stanford}  &  \multirow{2}{*}{4}    & 95.2M  &   $70.34^\uparrow$  & 8.15   & $25.03^\uparrow$ & $36.17^\uparrow$  & 34.92 & +4.02\\
	5 & ~~~ + Mixed Fine-tune~\cite{chu2017empirical} &  & 95.2M & 66.81 & 9.42 & 18.28 & $34.53^\uparrow$ & 32.26 & +1.36\\
	\hdashline
    6 & ~~~ + Domain Control~\cite{kobus2016domain} & \multirow{3}{*}{1} & 95.2M &	66.18 &	10.06 &	20.45 & 28.10	& 31.20 & +0.30\\
    7 & ~~~ + Domain Discriminate~\cite{britz2017effective} &  & 95.2M & 65.81	& 8.99 & 21.20 & $28.54^\uparrow$	& 31.15 & +0.25\\
	8   &   ~~~ + Domain Context~\cite{zeng2018multi} &     & 97.3M  & 66.81 & 9.75 & $22.74^\uparrow$ & $28.90^\uparrow$ & 32.05 & +1.15\\ 
	\hline
	9   &   ~~~ + Domain Transformation  & \multirow{2}{*}{1}  & 107.8M  & $67.70^\uparrow$ & 8.88 & $21.72^\uparrow$ & $31.07^\uparrow$ & 32.34 & +1.44 \\
	10   &   ~~~~~~ + Domain Supervision  & & 108.4M  &  66.63 & 8.23 & $28.43^\uparrow$ & $33.72^\uparrow$ & 34.25 & +3.35 \\
\end{tabular}
\caption{Translation results on small-scale {\em balanced} Zh$\Rightarrow$En multi-domain data used by Zeng et al. (2018). We also list the results of Zeng et al. (2018) on RNN-based NMT. ``\#M'' denotes the number of required models and ``\#Para.'' denotes the number of parameters. ``+'' denotes appending new features to the above row. ``$\uparrow$'' indicates statistically significant difference ($p < 0.01$) from ``Transformer'' in the corresponding domains.}  
\label{tab-results-zh-en}
\end{table*}

\paragraph{Data}

We conducted experiments on four different corpora, as listed in Table~\ref{tab-data}. For Chinese$\Rightarrow$English (Zh$\Rightarrow$En) translation, we used both a small-scale and a large-scale corpus. The small one is the same as that used by~\newcite{zeng2018multi}, and consists of four evenly distributed domains: \textit{law}, \textit{oral}, \textit{thesis} and \textit{news}. The large corpus is collected from CWMT2017 Lingosail, TVSub~\cite{wang2018translating} and LDC, which consists of four balanced domains: \textit{law}, \textit{news}, \textit{patent} and \textit{subtitle}. For German$\Rightarrow$English (De$\Rightarrow$En) and English$\Rightarrow$French (En$\Rightarrow$Fr) translation tasks, we used a large amount of training data extracted from OPUS. They respectively contain four and two {unevenly-distributed (unbalanced)} domains including \textit{law}, \textit{medical}, \textit{information technology} and \textit{Koran} and \textit{European Parliament}. The validation and test sets are officially-provided, otherwise randomly selected from the corresponding training corpora.

All the data were tokenized and then segmented into subword symbols using byte-pair encoding~\cite{sennrich2016subword} with 30K merge operations to alleviate the out-of-vocabulary problem. We used 4-gram BLEU score \cite{papineni2002bleu} as the evaluation metric, and bootstrap resampling~\cite{koehn2004statistical} for statistical significance.

\paragraph{Model}

For fair comparison, we implemented all proposed and other approaches on the advanced {\em Transformer} model~\cite{vaswani2017attention} using the open-source toolkit Fairseq~\cite{ott2019fairseq}.
We followed~\newcite{vaswani2017attention} to set the configurations of the NMT model, which consists of 6 stacked encoder/decoder layers with the layer size being 512. 
All the models were trained on 8 NVIDIA P40 GPUs where each was allocated with a batch size of 4,096 tokens. We trained the baseline model for 100K updates using Adam optimizer \cite{kingma2014adam}, and the proposed models were further trained with corresponding parameters initialized by the pre-trained baseline model. We fixed the hyperparameters $\lambda$ and $\delta$ as 0.1.

\paragraph{Baseline Comparisons} 
To make the evaluation convincing, we re-implemented and compared with five previous models on multi-domain adaptation, which can be divided into two categories with respect to the number of models. 
The multiple-model approaches require to maintain a dedicated NMT model for each domain:
\begin{itemize}

\item {\em Fine-tune}~\cite{luong2015stanford} that first trained a model on the entire data, and then fine-tuned multiple models separately using in-domain datasets.  

\item {\em Mixed Fine-tune}~\cite{chu2017empirical} that extended the fine-tune approach by training on out-of-domain data, then fine-tuning on in-domain and out-of-domain data.
\end{itemize}

\noindent The unified model methods handle adaptation to multiple domains within a unified NMT model:
\begin{itemize}
\item {\em Domain Control}~\cite{kobus2016domain} that introduced domain tag to the source sentence.
\item {\em Domain Discrimination}~\cite{britz2017effective} that adopted domain classification via multitask learning.
\item {\em Domain Context}~\cite{zeng2018multi} that incorporated the word-level context for domain discrimination.
\end{itemize}

\noindent Our work falls into the unified model, where the above three related approaches are comparable to ours.
Our work is not directly comparable to the fine-tuning approaches due to the different numbers of required models.

\begin{table*}[t]
\centering
\begin{tabular}{c|l||llll|lc}
	\multirow{2}{*}{\bf \#}    &  \multirow{2}{*}{\bf Architecture}   &    \multicolumn{6}{c}{\bf Zh$\Rightarrow$En}\\
	\cline{3-8}
	    & & Law & News & Patent & Tvsub & Avg. &{\bf  $\bigtriangleup$}\\
	\hline\hline
    1   &   Transformer & 38.77 & 49.05 & 47.68 & 30.30 & 41.45 & --\\
	\hline
	2   & ~~~ + Fine-tune~\cite{luong2015stanford}  & $42.32^\uparrow$ & $50.34^\uparrow$ & $49.16^\uparrow$ & 30.54 & 43.09 & +1.64 \\
    3   & ~~~ + Mixed Fine-tune~\cite{chu2017empirical} & $40.84^\uparrow$ & 49.53 & 46.45 & 30.95 & 41.91 & +0.49 \\
	\hdashline
    4   & ~~~ + Domain Control~\cite{kobus2016domain} & 39.27 & 49.30 & 48.02 & 30.55 & 41.79 & +0.34 \\
    5   & ~~~ + Domain Discriminate~\cite{britz2017effective} & 39.21 & 49.07 & 47.76 & 30.16 & 41.55 & +0.10 \\
    6   & ~~~ + Domain Context~\cite{zeng2018multi} & 39.35 & 49.77 & 47.71 & 30.31 & 41.79 & +0.34 \\
	\hline
	7   & ~~~ + Domain Transformation  & $40.04^\uparrow$ & $50.35^\uparrow$ & $48.35^\uparrow$ & 30.96 & 42.43 & +0.98  \\
	8   & ~~~~~~ + Domain Supervision  & $41.01^\uparrow$ & $50.55^\uparrow$ & $48.61^\uparrow$ & $31.55^\uparrow$ & 42.93 & +1.48 \\
\end{tabular}
\caption{Translation results on {\em large-scale} balanced Zh$\Rightarrow$En multi-domain data built in this work. ``+'' denotes appending new features to the above row. ``$\uparrow$'' indicates statistically significant difference ($p < 0.01$) from ``Transformer'' on different domains.}
\label{tab-results-zh-en-large}
\end{table*}

\begin{table*}[t]
\centering
\begin{tabular}{c|l||llll|lc||ll|lc}
	\multirow{2}{*}{\bf \#}    &  \multirow{2}{*}{\bf Architecture}   &    \multicolumn{6}{c||}{\bf De$\Rightarrow$En}   &    \multicolumn{4}{c}{\bf En$\Rightarrow$Fr}\\
	\cline{3-12}
	    & & Law & Med. & IT & Koran & Avg. &{\bf  $\bigtriangleup$} &   Med.  &   Par.    &   Avg. &{\bf  $\bigtriangleup$}\\
	\hline\hline
    1 &   Transformer      &   62.72    &   66.26   & 40.95 & 24.82  & 48.68 & -- &   65.91   &   35.58   &   50.75 & --\\
    \hline
	2 & ~~~ + Fine-tune  &   $65.10^\uparrow$  & $68.03^\uparrow$   & $42.76^\uparrow$ & 21.49 & 49.35   & +0.67 &  $68.56^\uparrow$   &   35.98   &  52.27 & +1.52 \\
    3 & ~~~ + Mixed Fine-tune & $63.48^\uparrow$ & 66.08 & 41.88 & $26.95^\uparrow$ & 49.60 & +0.92 & $67.87^\uparrow$ & 35.29 & 51.58 & +0.83 \\
    \hdashline
    4 & ~~~ + Domain Contr. & 63.04 & 66.69 & 41.13 & 24.27 & 48.78 & +0.10 & $67.01^\uparrow$  & 35.70 & 51.36 & +0.61 \\
    5 & ~~~ + Domain Discr. & 62.74 & 66.31 & 41.04 & 23.93 &	48.51 & -0.17 & 65.98 & 35.74 & 50.86 & +0.11 \\
    6 & ~~~ + Domain Conte. & 63.29 & 66.95 & 41.66 & 23.12 & 48.76	 & +0.08 & 66.57 & 35.67 & 51.12 & +0.37 \\
	\hline
	7 & ~~~ + Domain Trans.  &   63.33 & 66.95 & $42.32^\uparrow$ & 24.00 & 49.15  & +0.47 &  $67.23^\uparrow$ &   35.80   &   51.52 & +0.77 \\
	8 & ~~~~~~ + Domain Super.  & $64.59^\uparrow$ & $67.95^\uparrow$ & $42.16^\uparrow$ & 24.09 & 49.70 & +1.02 & $67.85^\uparrow$ & 35.80 & 51.83 & +1.08  \\
\end{tabular}
\caption{Translation results on {\em unbalanced} De$\Rightarrow$En and En$\Rightarrow$Fr multi-domain data. ``+'' denotes appending new features to the above row. ``$\uparrow$'' indicates statistically significant difference ($p < 0.01$) from ``Transformer'' on different domains.}
\label{tab-results-european}
\end{table*}

\subsection{Results}
\label{sec:4.3}

Table~\ref{tab-results-zh-en} and Table~\ref{tab-results-zh-en-large} respectively show results on the small-scale balanced Zh$\Rightarrow$En data used by~\newcite{zeng2018multi} and our newly-built large-scale corpus. Besides, Table~\ref{tab-results-european} shows results on Zh$\Rightarrow$En and Zh$\Rightarrow$En multi-domain data.
As seen, the proposed models significantly and incrementally improve the translation quality in all cases, although there are considerable differences among different scenarios.

\paragraph{Baselines} 
In Table~\ref{tab-results-zh-en}, the {Transformer} model (Row 3) greatly outperforms the results of RNN-based models reported by~\newcite{zeng2018multi} on the same data (Rows 1-2), which makes the evaluation convincing in this work.
The fine-tuning approaches (Rows 4-5) achieve significant improvements over the Transformer baseline. We attribute this to the facts that 1) the fine-tuning maintains a distinct model for each domain; and 2) there are sufficient data in each target domain. 
The unified models (Rows 6-8 in Table~\ref{tab-results-zh-en}) consistently improve translation performance, and the ``+Domain Context'' method achieves the best performance at the cost of introducing additional parameters. The unified models are directly comparable to our approach.

\paragraph{Our Models} As shown in Table~\ref{tab-results-zh-en}, the proposed models (Row 9-10) outperform not only the Transformer baseline (Row 3) but also comparable approaches (Rows 6-8). 
Introducing transformation networks (Row 9) improves translation performance over Transformer baseline by +1.44 BLEU point, indicating that \textsc{Dtn}s can effectively capture domain-aware knowledge.
Besides, adding two supervision signals (Row 10) can outperform the baseline by +3.35 BLEU. Surprisingly, the performance of our unified model is on par with fine-tuning which requires four separate models (34.25 vs. 34.92 BLEU). This is encouraging, since the fine-tune approach catastrophically increases the overhead of deployment in practice, while our approach avoids this problem without a significant decrease of translation performance.

\paragraph{Translation Quality on Other Scenarios}

To validate the robustness of our approach, we also conducted experiments on a large-scale Zh$\Rightarrow$En corpus (as shown in Table~\ref{tab-results-zh-en-large}) and other language pairs (as shown in Table~\ref{tab-results-european}). As seen, 
the superiority of our approach holds across different data sizes and language pairs, demonstrating the effectiveness and universality of the proposed approach.
Furthermore, our unified model surprisingly outperforms the fine-tuning (multiple-model) on the unbalanced De$\Rightarrow$En corpus.

\begin{figure}[t]
\centering
\includegraphics[width=0.44\textwidth]{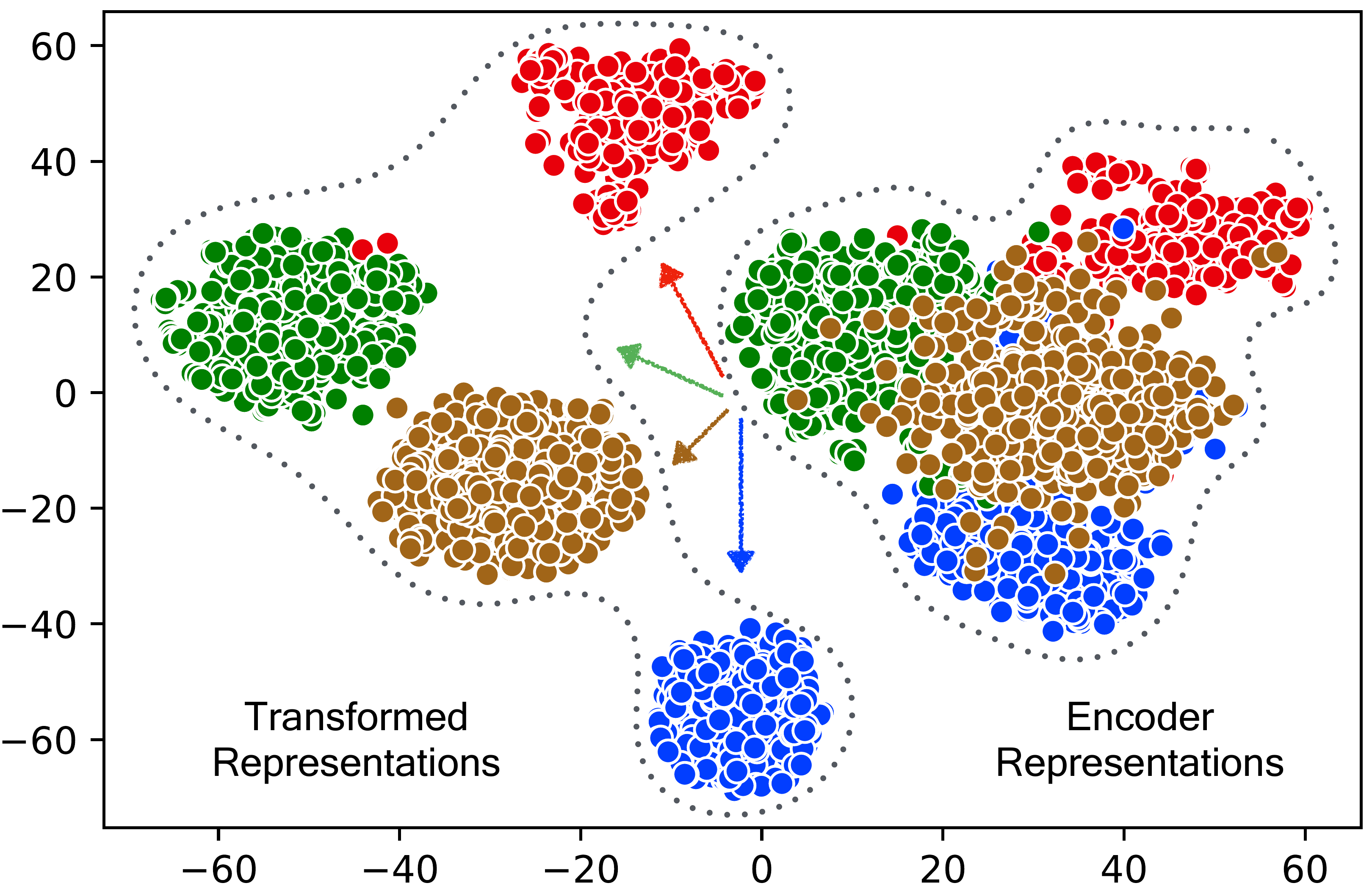}
\caption{Visualization of encoder (left) and transformed (right) representations. Dots in different colors denote sentences in different domains.}
\label{fig-tsne}
\end{figure}

\section{Analysis}

We conducted extensive analyses on the small-scale Zh$\Rightarrow$En data to better understand our model in terms of effectiveness of domain transformation and supervision.

\subsection{Effects of Domain Transformation}

\paragraph{Domain Transformation}
With the dimension reduction technique of t-SNE~\cite{maaten2008visualizing}, we visualized the general and domain-specific representations of source sentences in test set. As shown in Figure~\ref{fig-tsne}, the representation vectors in different domains are centered in different regions. Furthermore, the distribution of encoder representations is concentrated to preserve shared knowledge, while the transformed representations are diverse to keep domain-specific characteristics. This confirms that our approach is able to distinctively transform the source-side domain knowledge from the general to the particular.

\begin{figure}[t]
\centering
\includegraphics[width=0.43\textwidth]{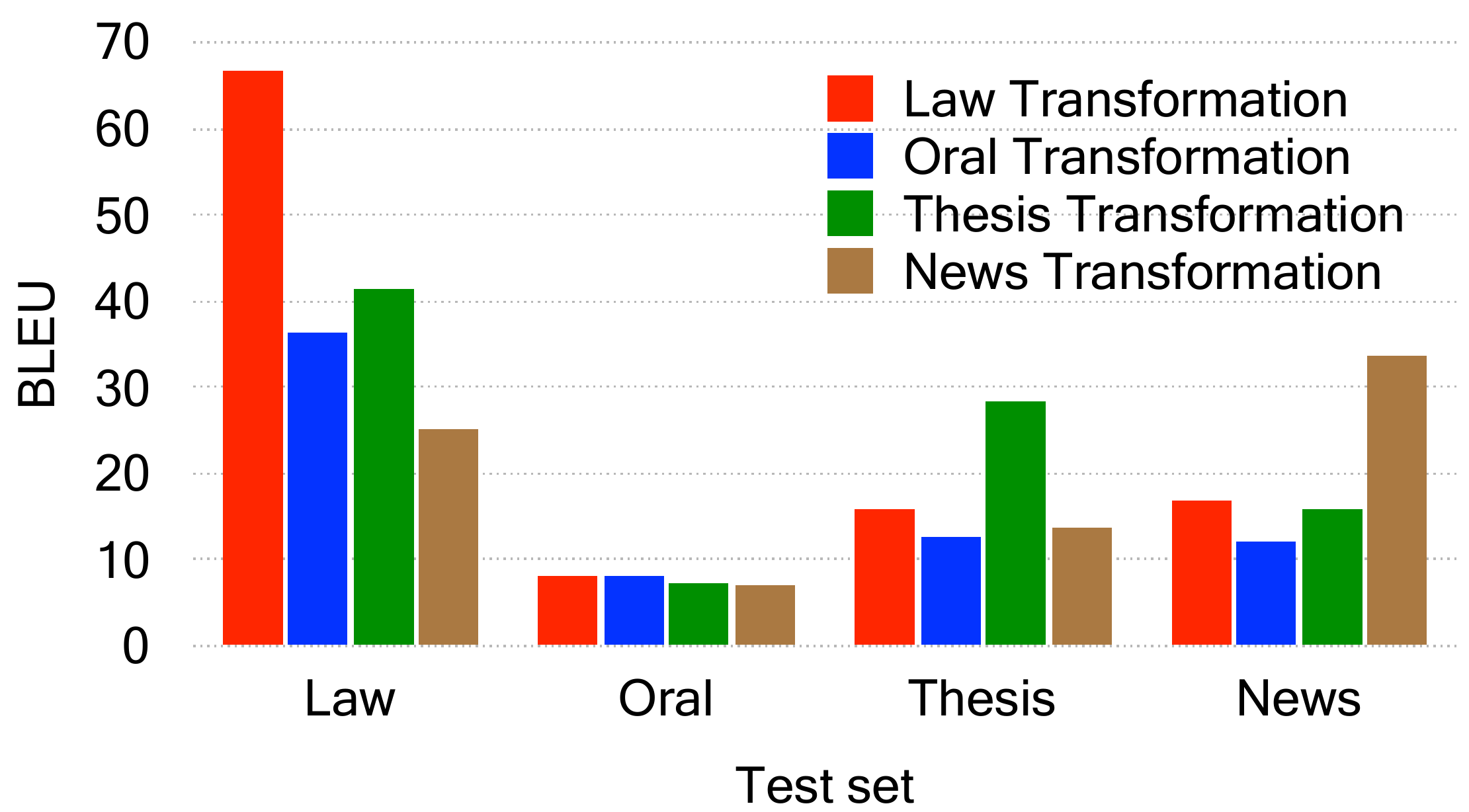}
\caption{Translation results of test set in each domain decoded by four domain-specific transformation modules. As seen, each specialized transformation model performs best on its corresponding domain.}
\label{fig-analyse-test}
\end{figure}

\paragraph{Domain-Specific Translation}
We further examined whether each specialized transformation module acquires its specific domain knowledge. Figure~\ref{fig-analyse-test} shows the translation results of test set in each domain decoded by four different domain-specific transformation modules. As seen, the each transformation module performs best on its corresponding domain. Some domains with more distinctive characteristics (e.g., \textit{Law}) can achieve more significant performances. In contrast, in less-distinctive domains (e.g., \textit{Oral}), different transformation modules have similar performances. This is consistent with our expectation that each transformation component is specialized to maintain particular knowledge in one domain.

\subsection{Effects of Domain Supervision}

\begin{table}[t]
  \centering
  \begin{tabular}{c|l|c}
      \bf \#  &  {\bf Model}  & {\bf BLEU} \\
      \hline \hline
      1 &   Transformer & 30.90\\
      2 &   ~~~ + Distillation~(sequence) & 31.45 \\
      3 &   ~~~ + Distillation~(word) & 31.51\\
    \hline
      4 &   ~~~ + Domain Transformation    &  32.34\\
      5 &   ~~~~~~ + Domain Distillation~(sequence) & 32.70\\
      6 &   ~~~~~~ + Domain Distillation~(word)  & 33.05\\
      7 &   ~~~~~~ + Domain Discrimination  & 33.18\\ 
      8 &   ~~~~~~~~~~ + Domain Distillation~(word) & 34.25\\
  \end{tabular}
  \caption{Translation results when different supervision signals are used for training our multi-domain model. ``Distillation (sequence)'' and ``Distillation (word)'' denote applying distillation at sequence and word level, respectively.} 
 \label{tab-supervision}
\end{table}

\paragraph{Contribution Analysis}

Table~\ref{tab-supervision} lists translation results when baseline or our model uses either {\em domain distillation} or {\em domain discrimination}, or both signals. As seen, adding supervision signal consistently improves the performance over the ``Domain Transformation'' model (Rows 6-7), and combining both signals accumulatively achieves the best performance (+1.9 BLEU, Row 8). This confirms the hypothesis in the section of domain supervision that the effects are reflected in three aspects: 1) weak supervision encourages model to exploit both shared and domain-aware knowledge across domains; 2) strong supervision guides model to learn distinct features; 3) combination makes them complementary to each other. 
It is also interesting to investigate the effect of domain supervision without transformation networks (Rows 1-3), which still improves performance (31.45 vs. 30.90), demonstrating the effectiveness and universality of domain supervision.

Concerning the distillation approach (Rows 2-3 and 5-6), we revisited word-level and sequence-level distillation methods for Transformer-based NMT. Different from the results reported by \newcite{kim2016sequence} on RNN-based models, we found that word-level distillation marginally outperformed its sequence-level counterpart (31.51 vs. 31.45 on top of ``Transformer'', and 33.05 vs. 32.70 on top of ``+ Domain Transformation''). Through case studies, we found that word-level distillation produced more fluent outputs, possibly due to providing smoother target labels. This explains why word-level distillation is a widely-used implementation in multi-lingual and multi-domain tasks on top of Transformer-based models~\cite{tan2019multilingual,you2019teach}. Therefore, we applied word-level distillation in our work.

\begin{table}[t]
\centering
\begin{tabular}{l|p{5.6cm}}
\hline
Input & 143 li yuan wai xin bo zhou ting huan zhe jing song lai yi yuan, fu su cun huo jin 2 li (1.4\%). \\
\hdashline
Reference & In the other 143 patients occurring \textcolor{blue}{sudden arrest of heart beat} \textcolor{red}{outside hospital}, only \textcolor{forestgreen}{2 survived} (1.4\%). \\
\hline\hline
Baseline & In the other 143 patients who received \textcolor{blue}{cardiac arrest}, only \textcolor{forestgreen}{2 survived} (1.4\%). \\ 
\hline
~~+Trans. & In the 143 patients admitted to \textcolor{red}{hospital}, only \textcolor{forestgreen}{2} (1.4\%) \textcolor{forestgreen}{survived} for \textcolor{blue}{resuscitation}.\\ \hdashline
~~~~+Distill. & In the other 143 patients who suffered \textcolor{blue}{a sudden arrest of heart beat} \textcolor{red}{outside hospital}, only \textcolor{forestgreen}{2} (1.4\%) \textcolor{forestgreen}{survived}.\\ 
\hdashline
~~~~~~+Discri. & In the other 143 patients who suffered \textcolor{blue}{sudden arrest of heart beat} \textcolor{red}{outside hospital}, only \textcolor{forestgreen}{2 survived} (1.4\%). \\
\hline
\end{tabular}
\caption{\label{table:example}
An example of Zh$\Rightarrow$En translation sampled from {\em Thesis} test set. Domain-specific words, phrases and patterns are highlighted with colors. Our ``+Trans.'', ``+Distill.'' and ``+Discri'' models are consistent with Table~\ref{tab-supervision}. As seen, augmenting transformation networks into NMT can generate more domain-specific words but with low fluency. Adding supervision signals can incrementally generate more fluent domain-specific phrases and patterns.}
\end{table}

\paragraph{Case Study}
Table~\ref{table:example} shows a translation example randomly selected from the test set in \textit{Thesis} domain. As seen, augmenting transformation module into NMT can generate more domain-specific words but with relatively lower fluency. Adding supervision signals can incrementally generate more fluent domain-specific phrases and patterns. For instance, the Chinese word ``yuan wai'' is ignored by baseline and mis-translated by ``+Trans.'' model, while the ``+Supervison'' models correctly translate it into ``outside hospital''. This demonstrates that our model can comprehensively capture domain-specific knowledge in terms of words, phrases and patterns.

\section{Related Work}
\paragraph{Domain Adaptation}
From conventional statistical machine translation (SMT) to state-of-the-art NMT, domain adaptation techniques have been widely investigated to adapt models trained on one or more source domains to outside target domain~\cite{chu2018survey,wang2017sentence,MarliesvanderWees17dynamic,chen2017cost,wang2017instance}. Although domain adaptation techniques boost translation quality on in-domain data, translation quality for out-of-domain data tends to degrade.

Fine-tune is the conventional way for domain adaptation~\cite{luong2015stanford,sennrich2015improving,freitag2016fast}.~\newcite{chu2017empirical} extended the fine-tune strategy by training the model on out-of-domain data, which is then fine-tuned on a mix of in-domain and out-of-domain data. The two approaches can be easily applied to multi-domain translation by separately maintaining a fine-tuned model for each domain. In this study, we empirically compare with the fine-tune strategies, and find that our unified model achieves comparable results with the fine-tuning approaches.

\paragraph{Multi-Domain Translation}
Multi-domain machine translation aims to construct the NMT model with the ability of translating sentences across different domains. \newcite{kobus2016domain} introduced embeddings of source domain tag to the encoder, which can perform domain-adapted translations in multiple domains. \newcite{britz2017effective} presented various mixing paradigms for multi-domain settings, and demonstrated their efficacy across multiple language pairs. \newcite{zeng2018multi} explored utilizing word-level domain contexts and jointly modeled multi-domain NMT and domain classification tasks. Our work is different in that 1) we learn the domain-specific knowledge by transforming from the general knowledge, while \newcite{zeng2018multi} split the encoder representation into general and domain-specific representations with two separate gates; and 2) we maintain a distinct transformation network with its own parameters for each domain, while \newcite{zeng2018multi} used a shared set of parameters across domains. In addition, we exploit more domain supervision techniques (e.g., domain distillation) to further improve multi-domain translation performance.

Furthermore, \newcite{gu2019improving} maintained a distinct set of encoder-decoder for each domain. This is analogous to the fine-tuning strategy, which maintains multiple models rather than a unified model for multi-domain translation. In addition, our approach also benefits from capturing the correlations between the general and domain-specific knowledge with the introduced transformation networks.

\section{Conclusion and Future Work}

In this paper, we propose to explicitly transform domain knowledge from the general to the particular for a multi-domain NMT model. In order to guarantee knowledge transformation, we also exploit two kinds of supervision signal to further improve the translation quality. Empirical results on a variety of language pairs demonstrate the effectiveness and universality of the proposed approach. We also conducted extensive analyses to demonstrate the necessity of explicitly modelling the transformation of domain knowledge for multi-domain translation. 

The proposed approach significantly improves translation performance at the cost of increased computational complexity. Network compression would be a promising direction to alleviate this problem. In future work we plan to exploit different model compacting techniques such as knowledge distillation~\cite{hinton2015distilling} and network pruning~\cite{han2015deep}, to make deployment of our approach more practical.

\bibliography{aaai}
\bibliographystyle{aaai}

\end{document}